\documentclass[11pt]{article}

\usepackage[top=2.5cm,left=2.54cm,right=2.54cm,bottom=2.5cm]{geometry}

\usepackage{natbib}
\usepackage{hyperref}

\usepackage{times}
\usepackage{latexsym}
\usepackage[T1]{fontenc}
\usepackage[utf8]{inputenc}
\usepackage{microtype}
\usepackage{inconsolata}
\usepackage{graphicx}
\usepackage{amsmath, amssymb}
\usepackage{booktabs}
\usepackage{caption}
\usepackage{subcaption}
\usepackage{array}
\usepackage{multirow}
\usepackage{xcolor}
\usepackage{CJKutf8}
\usepackage{tikz}
\usetikzlibrary{shapes.geometric, arrows.meta, positioning, fit, backgrounds}

\definecolor{vcol}{RGB}{15,110,86}
\definecolor{tcol}{RGB}{83,74,183}
\definecolor{acol}{RGB}{57,153,34}
\definecolor{scol}{RGB}{90,90,90}
\definecolor{lcol}{RGB}{185,60,30}
\definecolor{vbg}{RGB}{225,245,238}
\definecolor{tbg}{RGB}{238,237,254}
\definecolor{abg}{RGB}{234,243,222}
\definecolor{sbg}{RGB}{241,239,232}
\definecolor{lbg}{RGB}{250,236,231}

\tikzset{
  pbox/.style={
    rectangle, rounded corners=3pt, line width=0.5pt,
    text width=2.6cm, align=center,
    font=\scriptsize, inner sep=3pt, minimum height=0.75cm
  },
  vbox/.style={pbox, draw=vcol, fill=vbg},
  tbox/.style={pbox, draw=tcol, fill=tbg},
  auxbox/.style={pbox, draw=acol, fill=abg, text width=2.2cm},
  sbox/.style={pbox, draw=scol, fill=sbg, text width=3.0cm},
  lbox/.style={pbox, draw=lcol, fill=lbg},
  va/.style={-{Stealth[length=4pt]}, line width=0.8pt, color=vcol},
  ta/.style={-{Stealth[length=4pt]}, line width=0.8pt, color=tcol},
  ga/.style={-{Stealth[length=4pt]}, line width=0.8pt, color=scol},
  aux/.style={
    -{Stealth[length=3.5pt]}, line width=0.6pt,
    color=acol, dashed, dash pattern=on 3pt off 2pt
  },
}

\title{One Image, No Tokens: A Controlled Study of Glyph-Based Chinese Language Modeling}

\author{
Shuyang Xiang \\
\texttt{vanillaxiangshuyang@gmail.com} \\
\and
Hao Guan\thanks{Corresponding author.} \\
Tianli Qiming AI Research Institute, \\
Sichuan Qiming Daren Technology Co., Ltd., Chengdu, China \\
\texttt{guanhao032@gmail.com} \\
}

\begin{document}
\maketitle

\begin{abstract}
Modern language models generally represent text as sequences of discrete token
embeddings, an assumption deeply rooted in current practice but rarely
questioned. We challenge this representation, especially for Chinese, by
replacing index-based token embeddings entirely with a single rasterized image
of the character sequence, processed by a vision encoder composed of a shared
ResNet and a shallow Vision Transformer. To isolate the role of input
representation, we construct a dual-branch controlled framework in which both
a Vision-based model and an index-based baseline share an identical decoder
backbone, training objective, optimizer, and data curriculum. Any performance
difference is therefore attributable to the input modality only. Across all
tested decoder backbones, the Vision-based model consistently outperforms the
baseline, reaching a peak accuracy of 0.429 versus 0.355 for the index-based
baseline---a 21\% relative improvement---while converging in about half the
number of training epochs. The advantage emerges especially within the first
five epochs (under 21\% of total data) and persists under moderate character
corruption: the corrupted Vision model matches the \emph{clean} index-based
baseline. Ablation studies reveal that the advantage requires both spatially
coherent input and a ViT encoder with 2D positional encodings. A cross-script
comparison on English shows the advantage does not transfer directly to
alphabetic writing systems, suggesting that the uniform visual density and
radical structure of Chinese characters are enabling conditions. These findings
suggest that transformers are more modality-agnostic than commonly assumed, and
that discrete tokenization is not a fundamental requirement for Chinese language
modeling.
\end{abstract}

\section{Introduction}

Mainstream language models for logographic writing systems such as Chinese are
trained on discrete token embeddings, assuming that this index-based
representation is the natural input for modeling. In this work, we challenge
this convention by replacing index-based tokens entirely with a \emph{single}
rasterized image of the whole character sequence and train a model to predict
the next token purely from this visual input. We observe, perhaps surprisingly,
that this system learns next-token prediction effectively.

\begin{figure}[t]
\centering
\resizebox{\columnwidth}{!}{
\begin{tikzpicture}[node distance=0.45cm and 0.35cm]
\node[vbox] (vinput) {\textbf{Glyph image}\\rasterized sequence};
\node[vbox, below=of vinput] (resnet) {\textbf{ResNet}\\local features $\mathbf{z}_i$};
\node[vbox, below=of resnet] (transenc) {\textbf{ViT encoder}\\2D + 1D pos.\ enc.};
\node[vbox, below=of transenc] (proj) {\textbf{Linear proj.}\\patch $\to$ hidden dim};
\node[tbox, right=3.8cm of vinput] (tinput) {\textbf{Token IDs}\\discrete characters};
\node[tbox, below=of tinput] (temb) {\textbf{Token embedding}\\learnable table};
\node[tbox, below=of temb] (tpos) {\textbf{1D pos.\ enc.}\\learnable};
\node[tbox, below=of tpos] (tproj) {\textbf{Linear proj.}\\emb.\ $\to$ hidden dim};
\node[auxbox, right=0.6cm of resnet] (patchloss) {\textbf{Patch aux}\\$\times 0.3$};
\node[auxbox, right=0.6cm of transenc] (globalloss) {\textbf{Global aux}\\$\times 0.2$};
\node[sbox, below=1.0cm of proj, xshift=2.8cm] (gpt2) {\textbf{GPT-2 decoder}\\shared backbone};
\node[lbox, below=of gpt2] (mainloss) {\textbf{LM loss} $\times 1.0$};
\node[sbox, below=of mainloss] (output) {\textbf{Prediction}\\Acc@1 / Acc@5};
\draw[va] (vinput)--(resnet); \draw[va] (resnet)--(transenc);
\draw[va] (transenc)--(proj);
\draw[va] (proj.south)--++(0,-0.4)-|(gpt2.south west);
\draw[ta] (tinput)--(temb); \draw[ta] (temb)--(tpos);
\draw[ta] (tpos)--(tproj);
\draw[ta] (tproj.south)--++(0,-0.4)-|(gpt2.south east);
\draw[aux] (resnet.east)--(patchloss.west);
\draw[aux] (transenc.east)--(globalloss.west);
\draw[ga] (gpt2)--(mainloss); \draw[ga] (mainloss)--(output);
\node[above=0.12cm of vinput, font=\scriptsize\bfseries, color=vcol] {Vision branch};
\node[above=0.12cm of tinput, font=\scriptsize\bfseries, color=tcol] {Index-based branch};
\begin{pgfonlayer}{background}
  \node[draw=vcol!35,fill=vcol!5,rounded corners=5pt,
        fit=(vinput)(resnet)(transenc)(proj),inner sep=5pt,line width=0.4pt] {};
  \node[draw=tcol!35,fill=tcol!5,rounded corners=5pt,
        fit=(tinput)(temb)(tpos)(tproj),inner sep=5pt,line width=0.4pt] {};
\end{pgfonlayer}
\end{tikzpicture}
}
\caption{Dual-branch controlled pipeline. Vision branch renders the character
sequence as a glyph image; index-based branch uses discrete token IDs. Both
share the same GPT-2 decoder and LM objective. Auxiliary losses (dashed)
supervise the vision encoder only.}
\label{fig:pipeline}
\end{figure}
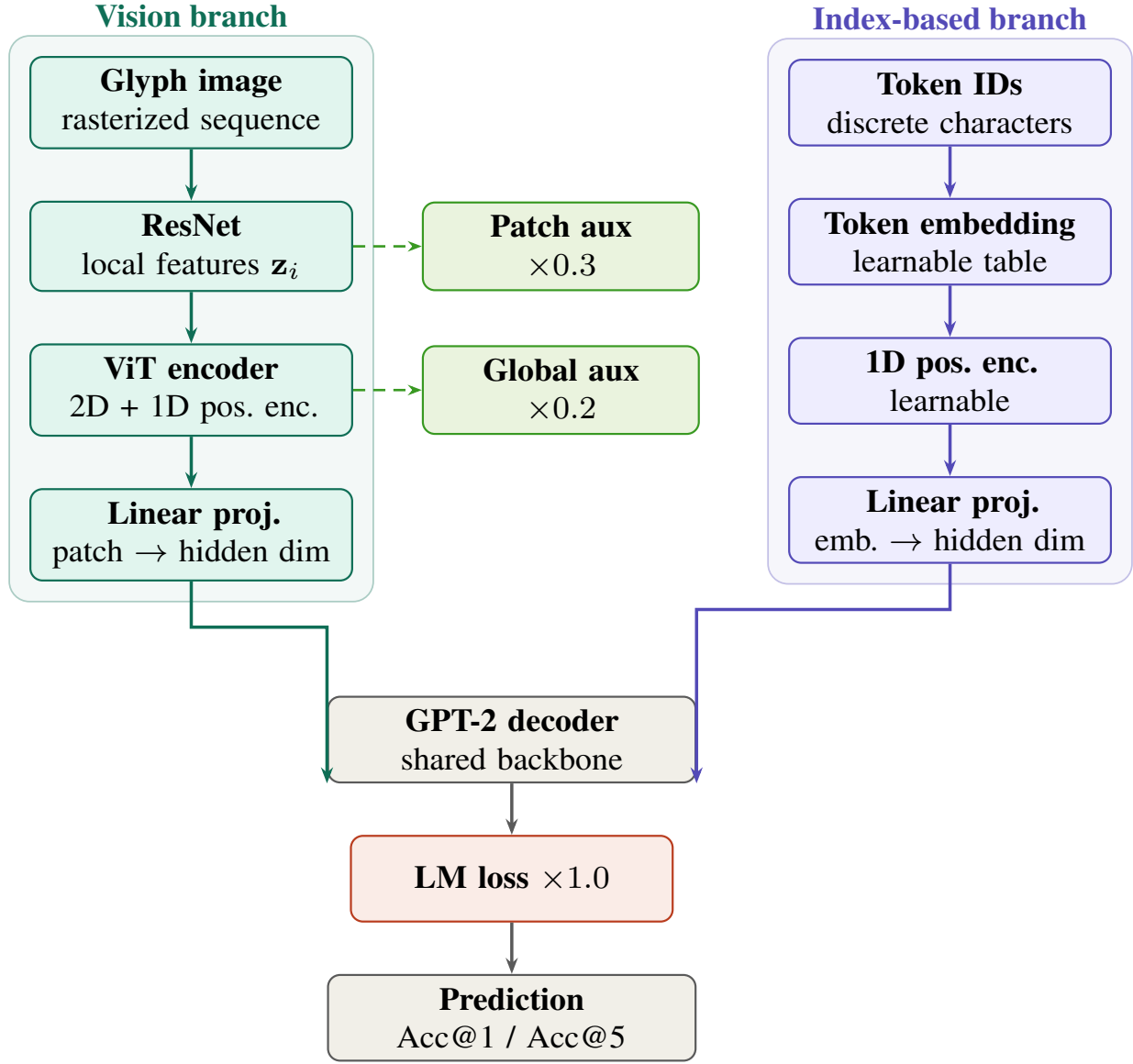

To rigorously study this phenomenon, we design a controlled dual-branch
framework with two models that share identical transformer backbones, main language-modeling
objectives, optimizers, schedulers, and data curriculum. The models differ in their input representations: one receives standard index-based token embeddings (index-based
baseline), while the other receives a single visual character image processed
by a vision encoder (Vision Model). The resulting performance differences therefore reflect the choice of input
representation and its associated encoding and training scheme, as illustrated in
Figure~\ref{fig:pipeline}.

\paragraph{Research Questions and Summary.}
We structure our studies around three main research questions:
\begin{itemize}
    \item \textbf{RQ1: Can vision-only representations outperform index-based
          token input for Chinese language modeling?} \\
          Yes. Across multiple backbone architectures, replacing index-based
          token embeddings with a vision-only input pipeline consistently
          improves next-token prediction performance, achieving up to a 21\%
          relative improvement in peak accuracy with GPT-2.
    \par\begingroup\samepage
    \item \textbf{RQ2: Do vision-based representations improve optimization
          dynamics during training?} \\
          Yes. The vision model converges faster, stops earlier (epoch~28
          vs.\ 47), and reaches its peak with fewer training samples---a
          compound advantage in both speed and sample efficiency.\par\endgroup
    \item \textbf{RQ3: Are vision-based representations more robust to input
          perturbations?} \\
          Conditionally. Vision-based inputs are more robust under moderate
          corruption (mask ratio $\leq 0.1$), but the index-based model's
          stronger discrete language prior becomes advantageous under heavy
          corruption (mask ratio $= 0.2$), revealing complementary robustness
          profiles for the two modalities.
\end{itemize}
To further understand these findings, we conduct multiple ablation studies and
additional analyses, including architectural variants, spatial perturbation
experiments, and downstream evaluation on C-Eval.
Visual representations of Chinese character sequences constitute a superior
input modality for language modeling under controlled conditions, outperforming
index-based token embeddings in sample efficiency, convergence speed, robustness
to corruption, and best observed validation accuracy across all decoder backbones tested.
Code is available at \url{https://github.com/ShuyangenFrance/chinese-vision}.
\section{Related Work}

\paragraph{Pixel-Based and Visual Language Models.}
Multiple prior works have explored representing text as images rather than
index-based tokens. \citet{broscheit2018learning} shows that pixel-derived
representations can match lookup embeddings in Chinese-to-English translation,
suggesting the potential of visual encoding for logographic scripts.
\citet{salesky2021robust} further demonstrate that rendering text as images
improves robustness to noise and out-of-vocabulary tokens.
PIXEL~\citep{rust2023pixel} and PIXAR~\citep{ai2024pixar} extend this idea to
masked and autoregressive image modeling on rendered text, showing strong
cross-lingual transfer. Multilingual extensions~\citep{kesen2025multilingual}
and CLIPPO~\citep{tschannen2023clippo} further explore unified visual-language
modeling without explicit tokenization.
A direct predecessor to the present work is \citet{xiang2025hotstart}, who
studied single-character visual rendering for Chinese language modeling using
the same GPT-2 backbone. That work found a pronounced early-training advantage
for visual inputs but observed that both modalities converge to essentially the
same final accuracy ($\sim$39\%). The present work substantially extends this
finding: by rendering the full sequence as a single 2D image rather than
treating each character independently, we observe not only a stronger
early-training advantage but also a persistent 21\% relative improvement in
peak performance, which we attribute to the spatial context encoded across
character positions.

\paragraph{Glyph and Visual Features for Chinese NLP.}
A related line of work incorporates glyph information as auxiliary signals in
token-based models. Glyce~\citep{Meng2019Glyce} introduces CNN-based glyph
embeddings using historical script variants. ChineseBERT~\citep{Sun2021ChineseBERT}
combines glyph and phonetic embeddings during pretraining. Subsequent work
explores radical-level supervision~\citep{zhu-etal-2023-glyph} and
sub-character visual properties in LLMs~\citep{Wu2025VisualImpact}.
LogogramNLP~\citep{DBLP:conf/acl/ChenSAMB24} benchmarks visual representations
for logographic scripts.

\paragraph{Vision Transformers as Sequence Encoders.}
Vision Transformers (ViT)~\citep{dosovitskiy2020vit} apply self-attention over
image patches and preserve spatial structure effectively~\citep{raghu2021vit}.
These properties make ViT-based encoders suitable for structured text images.

\paragraph{Language Modeling and Tokenization.}
Transformer language models such as GPT-2~\citep{radford2019language} rely on
discrete tokenization, which significantly affects performance, especially for
non-Latin scripts~\citep{rust-etal-2021-good}. Subword tokenization may
introduce representational biases~\citep{haslett-cai-2025-much}.

These observations motivate exploring input representations that replace
index-based tokenization entirely. Unlike prior work, which encodes text at
the token or character level or uses rendered text for generic pre-training
objectives, we study a vision-only framework that renders an entire Chinese
character sequence as a single image and studies the effect of this input pipeline under matched
decoder, data, and optimization settings.

\section{Methodology}
\label{sec:methodology}

\subsection{Problem Formulation}

Consider a sequence of characters $c_1, c_2, \dots, c_T$ and we work on the next-token
prediction task requiring $P(c_{t+1} \mid c_1, \dots, c_t)$.
We investigate whether this task can be learned when the model receives a
\textbf{single rasterized image of the entire character sequence} instead of a
sequence of discrete token IDs. All models are trained with the same language
modeling loss:
\[
\mathcal{L}_{LM} = \sum_{t=1}^{T}
  \text{CrossEntropy}\bigl(f(\mathbf{x}_{1:t}),\, c_{t+1}\bigr).
\]
We evaluate using top-$k$ accuracy Acc@$k$ and report Acc@1 and Acc@5. We adopt
accuracy rather than perplexity as our primary metric because model selection
and early stopping throughout this work are driven by top-$k$ next-character
prediction accuracy on the validation set, and because Acc@1/Acc@5 map directly
onto the quantity of practical interest for character-level decoding: whether
the correct character is produced (or ranked highly) at each step.
Peak validation accuracy is the highest accuracy among evaluated checkpoints,
not necessarily the accuracy at an epoch boundary. Perplexity
remains a standard secondary measure of the underlying probability
distribution and, where it aids comparison with prior tokenization work, we
report it alongside accuracy (e.g., Table~\ref{tab:disentangle}).

\subsection{Dual-Branch Controlled Framework}

\begin{itemize}
    \item \textbf{Index-based baseline}: characters as discrete token IDs,
          embedded via a learnable token embedding layer.
    \item \textbf{Vision Model}: the entire character sequence rasterized into
          a single 2D grayscale image, partitioned into non-overlapping
          $p{\times}p$ pixel patches, where $p$ is a configurable patch size.
\end{itemize}
Both representations feed into the \textbf{same pretrained decoder backbone}.
The main language-modeling objective, optimizer, learning rate schedule,
and curriculum are identical; the Vision model additionally uses auxiliary losses. Although the Vision model introduces
a ResNet and ViT, the total parameter count differs by only 0.4\% (113.7M
vs.\ 113.2M), providing a comparison at nearly matched total model sizes. Beyond this parameter-level control, the two model classes are also
architecturally symmetric: the index-based baseline and the Vision model share
the same GPT-2 decoder, the same output projection, and the same classification
head, differing only in how the input embedding for each position is produced.
Section~\ref{sec:encoder_ablations} isolates this further: Table~\ref{tab:encoder_ablations}
compares the encoder variants we test---including ablated versions with
substantially fewer effective parameters (\textit{single\_conv}) or no weight
sharing at all (\textit{no\_shared\_w}, $256\times$ more parameters than the
hybrid)---and shows that the hybrid outperforms the smaller single-layer
and larger unshared variants, suggesting that the advantage is not simply
a matter of extra encoder capacity.

\subsection{Data Preparation}
\label{sec:data_prep}

Both branches are trained on the same underlying character-level Chinese
corpus, so the same underlying text distribution is represented in two different
input modalities. We restrict the vocabulary to the 3,000
most frequent tokens in the training corpus---predominantly Chinese
characters, but also including a small number of digits and punctuation
symbols---plus two reserved special tokens, \texttt{<UNK>} (for tokens outside
the top-3,000) and \texttt{<EOS>}, giving a total vocabulary of 3,002 tokens;
tokens outside the top-3,000 are mapped to \texttt{<UNK>} for both branches
identically.

Training examples are constructed with a sliding window over the corpus: each
example is a 257-character span, of which the first 256 characters form the
context and the 257\textsuperscript{th} is the held-out prediction target
(Section~\ref{sec:methodology}). For the index-based baseline, the 256-character
context is tokenized into discrete IDs over the shared vocabulary. For the
Vision model, the same 256 characters are rasterized into a single glyph
image arranged as a $16{\times}16$ character grid and partitioned into
non-overlapping pixel patches (Section~\ref{sec:encoder_ablations}); the
target character is never rendered. Because both branches are built from
identical windows over identical corpus positions, the data curriculum
(Section~\ref{sec:methodology}) applies uniformly to both modalities.

\subsection{Vision Encoder}

\begin{figure*}[t]
\centering
\includegraphics[width=\textwidth]{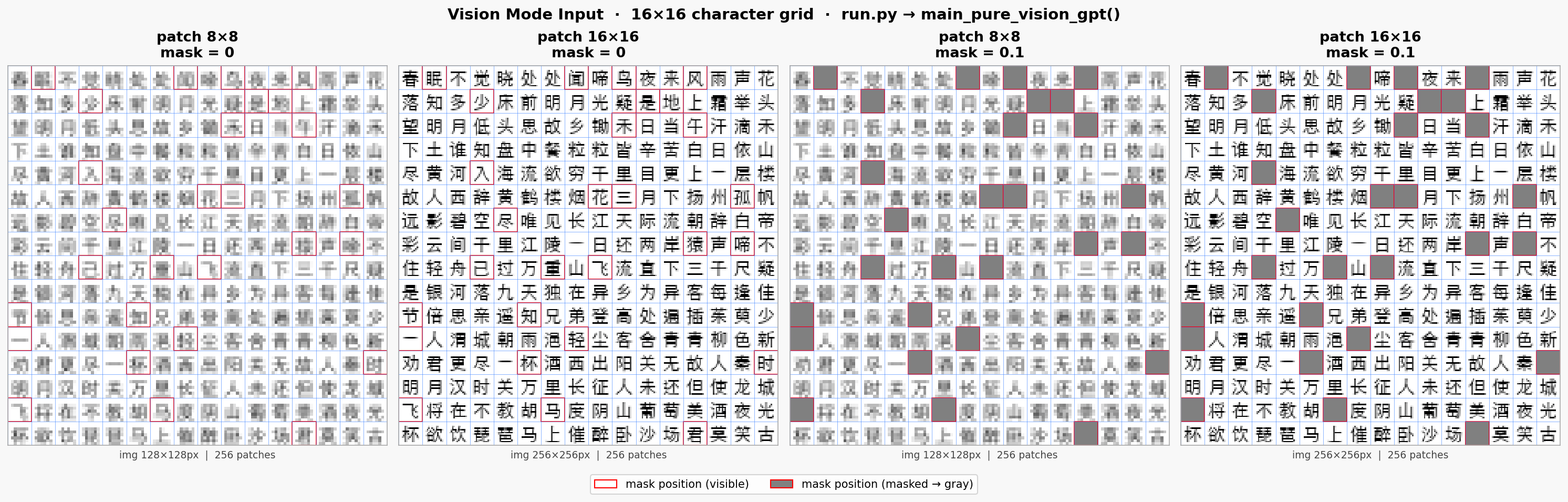}
\caption{Representative vision model inputs. Each panel shows a 16$\times$16
character grid (256 patches) rendered as a single grayscale image. Left two
panels: clean inputs at patch sizes $8{\times}8$ and $16{\times}16$. Right two
panels: the same inputs with 10\% random masking (gray fill, red border).}
\label{fig:patch_viz}
\end{figure*}

Figure~\ref{fig:patch_viz} illustrates representative vision model inputs under
different patch sizes and masking conditions.

A lightweight ResNet with shared weights extracts local glyph features:
$\mathbf{z}_i = f_{\text{ResNet}}(P_i)$.
These are augmented with learnable 2D and 1D positional encodings, then passed
through a ViT encoder~\citep{dosovitskiy2020vit}:
$\mathbf{V} = f_{\text{ViT}}(\{\mathbf{z}_i\})$.
The encoder outputs contextualized embeddings that directly replace discrete
index-based token embeddings.

\paragraph{Autoregressive causal constraint.}
Causality is enforced by construction of the training data rather than by an
additional mask inside the vision encoder. Each training example is built from
a fixed 257-character window: the first 256 characters form the prefix and are
rendered into the single glyph image (or, for the baseline, embedded as 256
discrete token IDs); the 257\textsuperscript{th} character is held out
exclusively as the prediction target and is never rendered into the image or
otherwise exposed to the encoder. Consequently, the visual input structurally
excludes any information about the character being predicted, regardless of
whether the ViT's cross-patch attention over the 256 given patches is itself
causally masked---there is nothing to leak, since the target patch does not
exist in the image. The pretrained GPT-2 decoder, which is architecturally
identical and shared across both branches, still applies its native
lower-triangular self-attention mask over the 256 context positions, exactly
as it would over an index-based token sequence. The autoregressive constraint
on the quantity of interest, $P(c_{257} \mid c_1,\dots,c_{256})$, therefore
holds identically for both modalities, and the ViT encoder's unmasked
attention operates only over already causally-valid context.

\subsection{Training Strategy}

Two auxiliary objectives supervise the vision encoder: (1) a patch-level
classification task ($\times 0.3$) requiring each patch to predict its depicted
character, and (2) a global classification objective ($\times 0.2$) predicting
the next token from the aggregated visual representation.
We adopt a quadratic curriculum:
$N_{\text{train}}(e) = N_{\text{begin}} + \alpha_2 e^2 + \alpha_1 e$,
with $N_{\text{begin}}=10{,}000$, $N_{\text{end}}=95{,}000$, $E=50$.
Both pipelines use AdamW with a 3-epoch linear warmup, OneCycle scheduling
(max lr $2\times10^{-4}$), weight decay 0.01, dropout 0.2, label smoothing 0.1.

\section{Experiments and Main Results}

We evaluate Vision models with patch size $p \in \{4, 8, 12, 16\}$ and mask
ratios $r \in \{0.05, 0.1, 0.2\}$. Validation is always on clean inputs.
Vision models use early stopping (patience 10): $8{\times}8$ stops at epoch~28,
$12{\times}12$ and $16{\times}16$ at epoch~31, $4{\times}4$ at epoch~37.
The index-based baseline trains for the full 50 epochs.

\subsection{Early Training Efficiency}
\label{sec:early}

\begin{figure}[t]
\centering
\includegraphics[width=\columnwidth]{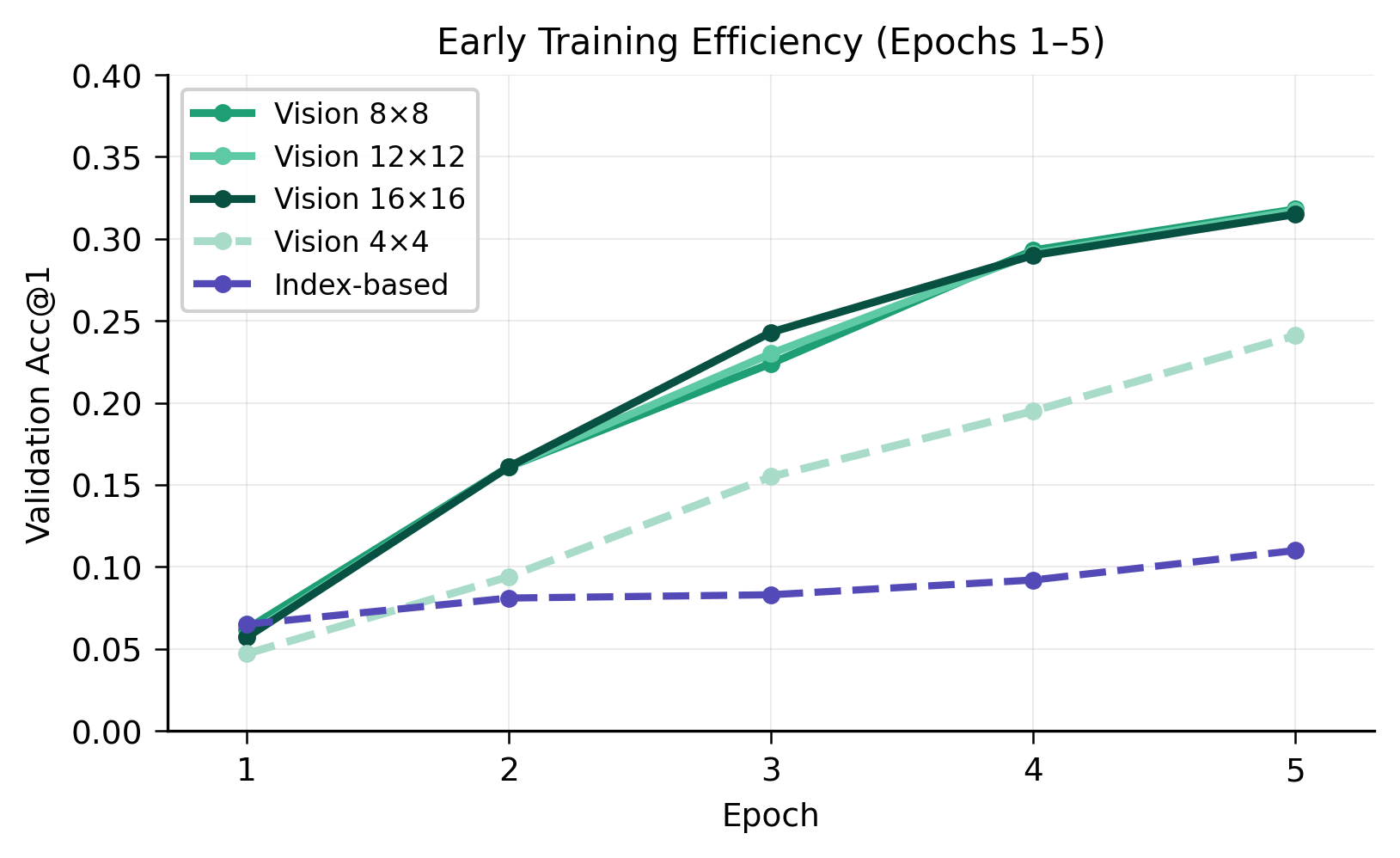}
\caption{Val Acc@1 during epochs 1--5. Vision $8{\times}8$, $12{\times}12$,
$16{\times}16$ cluster together from epoch~2; $4{\times}4$ lags due to
insufficient stroke detail.}
\label{fig:early_training}
\end{figure}

The first five epochs constitute the \emph{early training phase}, covering less
than 21\% of the final curriculum size. Table~\ref{tab:early} and
Figure~\ref{fig:early_training} show two phenomena.

\paragraph{Epoch-1 lag.}
Interestingly, at epoch~1, all Vision variants slightly trail the index-based
baseline (0.047--0.062 vs.\ 0.065). This lag is consistent across every patch
size and is absent from epoch~2 onward. We attribute it to an initialization
alignment cost: the ResNet and ViT encoder must first learn to produce
meaningful patch representations before they can usefully condition the
decoder, whereas the token embedding table can look up character identities
from the very first batch.

\paragraph{Hot-start advantage and patch-size interaction.}
From epoch~2, all Vision variants overtake the index-based baseline and the
gap widens. By epoch~5, the three larger variants ($8{\times}8$ through
$16{\times}16$) reach 0.315--0.318, about $2.9{\times}$ the accuracy of
the index-based baseline (0.110), while even the weakest variant ($4{\times}4$,
0.241) is about $2.2{\times}$ as accurate. The $4{\times}4$ variant lags at
epoch~2 (0.094 vs.\ 0.161), which we attribute to a resolution floor:
$4{\times}4$ pixels per character provides insufficient stroke detail.

\begin{table}[t]
\centering\small
\begin{tabular}{c|c|cccc}
\hline
Ep & Index & \multicolumn{4}{c}{Vision (patch px)} \\
   &       & $4{\times}4$ & $8{\times}8$ & $12{\times}12$ & $16{\times}16$ \\
\hline
1 & \textbf{.065} & .047 & .062 & .058 & .057 \\
2 & .081 & .094 & \textbf{.161} & \textbf{.161} & \textbf{.161} \\
3 & .083 & .155 & .224 & .230 & \textbf{.243} \\
4 & .092 & .195 & \textbf{.293} & .291 & .290 \\
5 & .110 & .241 & \textbf{.318} & .317 & .315 \\
\hline
\end{tabular}
\caption{Val Acc@1 during the early training phase (epochs 1--5).}
\label{tab:early}
\end{table}

\subsection{Robustness Under Character Corruption}

The same character positions are masked in both modalities. All corruption
experiments use the $8{\times}8$ patch configuration; validation is always on
clean inputs.

\paragraph{Moderate corruption (mask $= 0.1$).}
Table~\ref{tab:corrupt_learning} reports full training dynamics under 10\%
corruption. When a token is corrupted in the index-based baseline, its discrete
identity is destroyed entirely. In the Vision model, corruption replaces a patch
with a gray fill, but the surrounding 90\% of patches retain their full spatial
layout and stroke detail. Notably, the Vision model \emph{with} 10\% corruption
(peak 0.356) still matches the clean index-based baseline (peak 0.355).

\begin{table}[t]
\centering\small
\begin{tabular}{c|cc|cc}
\hline
Ep & Index @1 & Index @5 & Vision @1 & Vision @5 \\
\hline
1  & .040 & .170 & .016 & .141 \\
4  & .075 & .290 & .213 & .406 \\
8  & .162 & .355 & .288 & .477 \\
12 & .197 & .396 & .311 & .508 \\
20 & .252 & .430 & .341 & .537 \\
24 & .262 & .443 & \textbf{.356} & \textbf{.559} \\
47 & \textbf{.307} & \textbf{.481} & --- & --- \\
\hline
\end{tabular}
\caption{Val accuracy under 10\% training-time corruption ($8{\times}8$).
Vision stops at epoch~24; index-based baseline trains for 47 epochs.}
\label{tab:corrupt_learning}
\end{table}

\paragraph{Effect of mask ratio.}
Table~\ref{tab:mask_ratio} reveals a striking asymmetry. As the mask ratio
increases from 0.05 to 0.20, Vision Acc@1 falls from 0.331 to
0.209---a 37\% drop relative to its performance at mask ratio 0.05---while the index-based
model falls only modestly (0.302 to 0.267). At mask ratio 0.20, the
relationship inverts entirely ($-21.7\%$ for Vision). This reflects stronger
n-gram co-occurrence priors in discrete tokens under heavy corruption; Vision's
out-of-distribution gray fills degrade encoder learning beyond a threshold.
A diagnostic: validation accuracy exceeds training accuracy in all masked runs
(masking acts as training augmentation), and Vision outperforms the index-based baseline under light
augmentation (mask $= 0.05$: $+9.6\%$) but has a narrower tolerance window.

\begin{table}[t]
\centering\small
\setlength{\tabcolsep}{5pt}
\begin{tabular}{c|cc|c}
\hline
Mask & Vision Acc@1 & Index Acc@1 & $\Delta$ \\
\hline
0.00 (clean) & .369 & --- & --- \\
0.05 & .331 & .302 & $+9.6\%$ \\
0.10 & .290 & .264 & $+9.8\%$ \\
0.20 & .209 & .267 & $-21.7\%$ \\
\hline
\end{tabular}
\caption{Val Acc@1 at epoch~10 under different mask ratios. Val on clean inputs.}
\label{tab:mask_ratio}
\end{table}

\subsection{Full Training Dynamics}

\begin{figure}[t]
\centering
\includegraphics[width=\columnwidth]{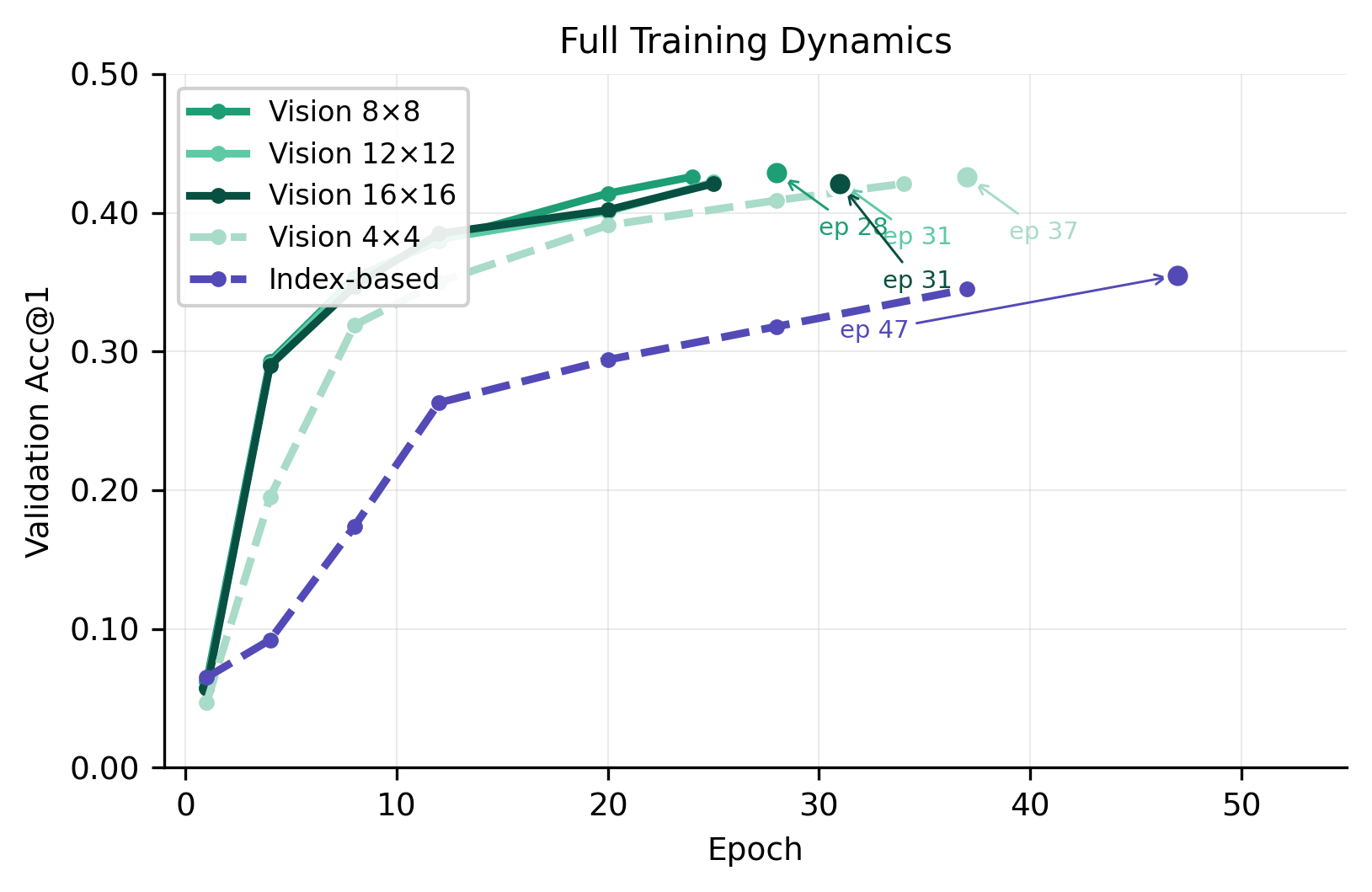}
\caption{Full training dynamics. Circle markers indicate early stopping epochs.}
\label{fig:full_training}
\end{figure}

Table~\ref{tab:patch_size_full} and Figure~\ref{fig:full_training} show that
every Vision variant substantially outperforms the index-based baseline.
The $8{\times}8$ configuration achieves the highest peak (0.429 at epoch~28)
versus 0.355 for the index-based baseline at epoch~47---a 21\% relative
improvement in substantially fewer training epochs. The $4{\times}4$ variant
ultimately converges to a competitive peak (0.426 at epoch~37), showing that
even low-resolution patches encode enough character information given sufficient
training, but at the cost of slower optimization.

\begin{table}[t]
\centering\small
\begin{tabular}{l|cc|c}
\hline
Model & Peak Acc@1 & Peak Acc@5 & Stop \\
\hline
Index-based & 0.355 & 0.529 & 47 \\
\hline
Vision $4{\times}4$   & 0.426 & 0.626 & 37 \\
Vision $8{\times}8$   & \textbf{0.429} & \textbf{0.629} & 28 \\
Vision $12{\times}12$ & 0.422 & 0.622 & 31 \\
Vision $16{\times}16$ & 0.421 & 0.621 & 31 \\
\hline
\end{tabular}
\caption{Peak validation accuracy and early stopping epoch, clean conditions.}
\label{tab:patch_size_full}
\end{table}

\subsection*{Summary of Main Results}

Three consistent findings: (1) Vision exhibits a pronounced hot-start advantage
from epoch~2, robust to patch size above a resolution floor. (2) Complementary
robustness profiles: Vision leads under light corruption (mask $\leq 0.1$),
while the discrete prior prevails under heavy corruption (mask $= 0.2$). (3)
Vision converges faster and to higher peaks across all patch sizes.

\section{Analysis and Ablation Studies}
\label{sec:ablation}

All ablations use 10 epochs due to resource constraints.

\subsection{Encoder Component Ablations}
\label{sec:encoder_ablations}

We examine four architectural variants: \textbf{single\_conv} (single-layer
convolutional encoder), \textbf{no\_shared\_weights} (256 independent encoders),
\textbf{ResNet-only} (bypassing the Transformer), and \textbf{ViT-only}
(removing the ResNet, linear projection to ViT).

Table~\ref{tab:encoder_ablations} shows that the hybrid achieves the highest
reported accuracy among the tested visual encoder variants. The hybrid model (0.369) outperforms both ResNet-only
(0.304) and ViT-only (0.338): ResNet provides rich local glyph features while
ViT aggregates them across the 2D spatial layout. At epoch~1, ResNet-only and
ViT-only start higher (0.096--0.104 vs.\ 0.062), suggesting the two-stage
encoder requires a brief alignment period. Simplifying the ResNet to a single
layer retains much of the hybrid model's performance; removing weight sharing
hurts more despite $256{\times}$ more parameters, confirming
position-invariant weight sharing as a beneficial inductive bias.

\begin{table}[t]
\centering\small
\setlength{\tabcolsep}{5pt}
\begin{tabular}{l|cc|cc|cc}
\hline
Model & @1 & @5 & @1 & @5 & @1 & @5 \\
      & ep1 & ep1 & ep4 & ep4 & ep10 & ep10 \\
\hline
Vision hybrid & .062 & .169 & .293 & .481 & \textbf{.369} & \textbf{.567} \\
ViT-only      & .096 & .229 & .197 & .399 & .338 & .546 \\
ResNet-only   & .104 & .238 & .195 & .402 & .304 & .518 \\
single\_conv  & .051 & .134 & .223 & .397 & .350 & .550 \\
no\_shared\_w & .042 & .106 & .183 & .346 & .313 & .506 \\
Index-based   & .065 & .209 & .092 & .345 & ---  & ---  \\
\hline
\end{tabular}
\caption{Encoder ablations at epochs 1, 4, 10 (Acc@1 and Acc@5).}
\label{tab:encoder_ablations}
\end{table}

\subsection{Disentangling Spatial Layout and Contextual Encoding}
\label{sec:disentangle}

The encoder ablations establish that the hybrid outperforms its components, but
do not explain \emph{why}. We conduct a controlled three-condition experiment
with a frozen GPT-2 decoder and precomputed PatchResNet features, so the only
variable is the positional encoding and contextual encoder:
\textbf{A} (1D order embedding), \textbf{B} (2D-aware ResNet, no Transformer),
\textbf{C} (2D positional encoding + ViT, 2 layers, 8 heads).

\begin{table}[t]
\centering\small
\begin{tabular}{cl|c|c|c}
\hline
 & Condition & Acc@1 & Acc@5 & PPL \\
\hline
A & 1D baseline    & 9.35\% & 20.49\% & 589.22 \\
B & 2D ResNet-only & 9.56\% & 21.26\% & 580.01 \\
C & 2D Full        & \textbf{12.00\%} & 25.29\% & 476.93 \\
\hline
\end{tabular}
\caption{Disentanglement experiment. All conditions use the same frozen
GPT-2 and precomputed features.}
\label{tab:disentangle}
\end{table}

The A$\to$B gap is negligible ($+$0.21 pp, PPL $-$9.2); the B$\to$C gap is
substantial ($+$2.44 pp, PPL $-$103.1). Spatial layout is \emph{necessary} but
not \emph{sufficient}: the ViT's cross-patch attention guided by 2D positional
encodings converts spatial structure into language-predictive representations.

\subsection{Generalization Across Backbones}
\label{sec:backbones}

\begin{table}[t]
\centering\small
\begin{tabular}{c|cc|cc}
\hline
Ep & Index @1 & Index @5 & Vision @1 & Vision @5 \\
\hline
1  & .079 & .246 & .075 & .195 \\
2  & .144 & .365 & .244 & .430 \\
4  & .238 & .445 & .331 & .528 \\
6  & .284 & .474 & .369 & .567 \\
8  & .290 & .477 & .383 & .586 \\
10 & .281 & .451 & \textbf{.399} & \textbf{.597} \\
\hline
\end{tabular}
\caption{Validation accuracy with Qwen2.5-0.5B decoder. Index-based baseline
shows a late decline in validation accuracy, suggesting overfitting.}
\label{tab:qwen25}
\end{table}

\begin{table}[t]
\centering\small
\begin{tabular}{c|cc|cc}
\hline
Ep & Index @1 & Index @5 & Vision @1 & Vision @5 \\
\hline
1  & .148 & .358 & .154 & .321 \\
2  & .202 & .431 & .273 & .459 \\
4  & .258 & .479 & .351 & .547 \\
6  & .309 & .493 & .382 & .584 \\
8  & .312 & .494 & .395 & .595 \\
10 & .338 & .517 & \textbf{.395} & \textbf{.597} \\
\hline
\end{tabular}
\caption{Validation accuracy with Qwen3.5-0.8B decoder. Training Acc@1
reaches 0.885 by epoch~10 while validation accuracy remains much lower, suggesting overfitting
in the index-based baseline.}
\label{tab:qwen35}
\end{table}

\begin{table}[t]
\centering\small
\begin{tabular}{l|cc|c}
\hline
Backbone & Vision & Index & Gain \\
\hline
GPT-2 (ep 28/47)     & \textbf{.429} & .355 & $+20.8\%$ \\
Qwen2.5-0.5B (ep 10) & \textbf{.399} & .281 & $+42.0\%$ \\
Qwen3.5-0.8B (ep 10) & \textbf{.395} & .338 & $+16.9\%$ \\
\hline
\end{tabular}
\caption{Vision vs.\ Index Acc@1 across three decoder backbones.}
\label{tab:backbone_summary}
\end{table}

The vision advantage holds across all three backbones (Table~\ref{tab:backbone_summary}).
The Qwen baselines show signs of overfitting: validation accuracy declines
late in training for Qwen2.5, while Qwen3.5 has a large training--validation
gap. Vision achieves higher validation accuracy with both backbones,
consistent with better generalization under the curriculum's growing data regime. The early training advantage from Section~\ref{sec:early} replicates
across backbones: Vision reaches 0.244--0.273 by epoch~2 while index-based
is at 0.144--0.202, suggesting that the visual input pipeline provides an inductive bias
that transfers across the tested backbones.

\subsection*{Summary of Ablation Studies}

The hybrid ResNet+ViT encoder achieves the best reported performance
among the tested encoder variants; weight sharing
is a key inductive bias; 2D spatial layout and ViT cross-patch attention work best in combination
among the tested configurations; and the advantage generalizes across three
decoder backbones.

\section{Discussion}

\subsection{Why Does the Visual Advantage Arise?}

Our experiments point to complementary roles for spatial layout and
contextual encoding. The disentanglement experiment
(Section~\ref{sec:disentangle}) shows that spatial layout alone offers only a
small gain: a 2D-aware ResNet without a spatial Transformer
encoder yields negligible improvement ($+$0.21 pp) over the 1D baseline,
whereas adding the ViT with 2D positional encodings produces a substantial gain
($+$2.44 pp). Together, these findings point to a specific mechanism: the ViT
encoder, guided by 2D positional encodings, learns to aggregate stroke and
radical information across spatially coherent character patches into
language-discriminative representations. Discrete token embeddings, by contrast,
must accumulate this structural knowledge purely through co-occurrence
statistics, requiring more data and more training epochs to reach comparable
performance. This also explains the epoch-1 lag observed in
Section~\ref{sec:early}: the two-stage vision encoder requires an initial
alignment period before its representations become useful to the decoder,
whereas token embeddings can immediately look up character identities.

\subsection{Embedding Update Frequency and Weight Tying}
\label{sec:update_frequency}

A related question is whether the index-based baseline is disadvantaged by
uneven gradient updates: rare characters appear infrequently in training, so
their rows in the token embedding table receive comparatively few gradient
updates, whereas the Vision model's ResNet and ViT weights are shared across
all character identities (Section~\ref{sec:encoder_ablations}) and therefore
receive a gradient contribution from \emph{every} character occurrence in
training, regardless of that character's own frequency. We view this update-frequency asymmetry as a plausible contributor to
the advantage of shared visual encoders. The encoder
ablation in Table~\ref{tab:encoder_ablations} provides related evidence: removing
weight sharing (\textit{no\_shared\_w}, 256 independent per-position encoders)
substantially lowers Vision performance despite a $256\times$ increase in
parameters, suggesting that shared-weight structure---potentially through more frequent
updates that also benefit rare characters---contributes to the advantage,
rather than raw encoder size alone. Both branches otherwise receive
identical treatment: the index-based baseline's token embedding table is
updated with standard per-batch AdamW gradients with no custom frequency
scheduling, and the vocabulary (top-3,000 tokens plus
\texttt{<UNK>} and \texttt{<EOS>}; Section~\ref{sec:data_prep}) is shared across branches. We do not use weight
tying between the input embedding and output classification head in either
branch; the two are independently parameterized. We did not evaluate an
index-based baseline using decomposed sub-character units (e.g., radical or
stroke tokenization), which could in principle narrow this update-frequency
gap by sharing statistics across characters that share components; we leave
this comparison to future work.

\subsection{Downstream Transfer (C-Eval)}

We evaluate downstream transfer on C-Eval~\citep{huang2023ceval}, a Chinese
multiple-choice knowledge and reasoning benchmark, using the official
validation split (\texttt{ceval/ceval-exam}, \texttt{val}, which includes
answer keys). Evaluation is zero-shot: for each question we construct a
prompt containing the question stem followed by its four labeled options
(A/B/C/D), format it identically to the training-time input for each branch
(discrete character IDs for the index-based baseline; the same prompt
rasterized into a glyph grid for the Vision model), and take the model's
final-position output logits. The predicted answer is the option letter
(A, B, C, or D) with the highest log-probability at that position; this is a
pure scoring procedure with no free-form generation or text matching involved.
Table~\ref{tab:ceval} reports six representative subjects from C-Eval; each
subject's val split contains more questions than we evaluate here, and we
report results on a subset of $n\!\approx\!20$--$22$ questions per subject
(130 questions total across the six subjects shown), rather than the full
validation set, due to the small-scale nature of the underlying language
models. We flag this explicitly as a limitation of the downstream evaluation:
results at this sample size are indicative rather than statistically robust,
and should not be read as C-Eval leaderboard-comparable scores.

\begin{table}[t]
\centering\small
\setlength{\tabcolsep}{5pt}
\begin{tabular}{l|cc|cc}
\hline
Subject & V(best) & I(best) & V(ep10) & I(ep10) \\
\hline
HS Biology   & \textbf{42.1} & 10.5 & 42.1 & 42.1 \\
HS Geography & \textbf{26.3} & 21.1 & 21.1 & 21.1 \\
HS History   & \textbf{35.0} & 25.0 & 30.0 & 30.0 \\
HS Politics  & 21.1 & \textbf{26.3} & 21.1 & 21.1 \\
Edu. Science & \textbf{24.1} & 20.7 & 24.1 & 27.6 \\
Law          & \textbf{29.2} & 16.7 & 20.8 & 20.8 \\
\hline
\textbf{Overall} & \textbf{27.7} & 20.0 & 26.2 & 26.9 \\
\hline
\end{tabular}
\caption{C-Eval downstream evaluation (\%). V=Vision, I=Index. Random
chance is 25\%.}
\label{tab:ceval}
\end{table}

Vision achieves 27.7\% overall versus 20.0\% for the index-based baseline
at best-checkpoint evaluation (Table~\ref{tab:ceval}), suggesting the visual
inductive bias partially transfers to knowledge-intensive tasks. The advantage
is consistent across five of six subjects; High School Politics is the
exception, possibly due to its reliance on memorized factual associations.

\subsection{Is the Advantage Specific to Chinese?}

\begin{table}[t]
\centering\small
\begin{tabular}{c|cc|c}
\hline
Ep & Vision $8{\times}8$ & Vision $16{\times}16$ & Index \\
\hline
1  & .114 & .112 & .146 \\
4  & .177 & .169 & .215 \\
6  & .200 & .199 & .240 \\
8  & .220 & .215 & .265 \\
10 & .229 & .226 & .291 \\
\hline
\end{tabular}
\caption{English experiment (WikiText-2). Unlike Chinese, the index-based
baseline leads throughout.}
\label{tab:english}
\end{table}

On WikiText-2 (English, Table~\ref{tab:english}), the index-based baseline
leads from epoch~1 and the gap widens overall---a complete reversal of
the Chinese finding. This contrast is not surprising given a structural
asymmetry between the two writing systems. In Chinese, each character occupies
a fixed visual unit with consistent spatial extent and rich internal structure
(strokes, radicals) that encodes partial semantic information. In English,
subword tokens vary arbitrarily in length and visual complexity: a single-letter
token such as ``a'' and a multi-character token such as ``international'' are
both assigned the same fixed patch size, producing representations of
incomparable visual density. This asymmetry has no counterpart in Chinese,
where the one-character-one-patch correspondence is exact and uniform.
These results suggest that the visual advantage depends on the writing
system and rendering strategy, with Chinese glyphs providing compact,
information-dense visual units.

\paragraph{Possible remedies for alphabetic rendering.}
Several alternative rendering strategies for English could in principle reduce
the visual density mismatch, but each introduces its own problems.

\textit{Horizontal-only scaling}: one could keep the vertical dimension fixed
at 8 pixels while scaling the horizontal dimension proportionally to token
length. This preserves letter size uniformity but makes patch width a direct
encoding of token length, giving the model a trivial structural cue unrelated
to visual semantics. More importantly, it destroys morphological consistency:
the prefix \textit{inter-} would appear at a different horizontal scale in
``international'' versus ``internet'', preventing the encoder from learning
shared visual representations for shared morphemes.

\textit{Proportional fixed-height rendering}: an alternative is to render every
token at a fixed letter size within a variable-width patch. This preserves
letter size but requires variable-width patches, incompatible with the
fixed-grid architecture used here. It would also assign vastly different amounts
of visual real estate to common short tokens versus rare long tokens.

\textit{Character-level rendering}: a third option is to abandon subword
tokenization entirely and render individual letters as patches, giving a uniform
one-letter-one-patch correspondence analogous to Chinese. However, the Latin
alphabet contains only 26 base letters, offering far less visual diversity than
thousands of Chinese characters. The fundamental limitation is not the rendering
strategy but the \emph{semantic density} of the visual unit: Chinese characters
carry stroke-level and radical-level semantic cues that have no counterpart in
Latin letters.

These considerations suggest that a principled extension of glyph-based
language modeling to alphabetic scripts would require either a fundamentally
different tokenization unit or a writing system that, like Chinese, encodes
partial semantics in the visual form of each token.

\section{Conclusion}

We showed that rendering Chinese character sequences as single glyph images and
processing with a ResNet+ViT encoder consistently outperforms index-based token
embeddings under controlled conditions: 21\% higher peak accuracy, faster
convergence, better sample efficiency, and complementary robustness profiles
under corruption. The advantage requires both spatially coherent input and ViT
cross-patch attention with 2D positional encodings, and does not transfer to
the tested English rendering configuration, suggesting a role for the uniform
visual density of Chinese characters. These findings suggest that transformers are more
modality-agnostic than commonly assumed, and motivate closer examination of how
writing system properties interact with neural sequence model inductive biases.

\section*{Limitations}

\textbf{Scale.} Experiments use decoders up to 868M parameters and a
3,002-token vocabulary consisting of the 3,000 most frequent tokens in the
training corpus (predominantly Chinese characters, with a small number of
digits and punctuation symbols) plus \texttt{<UNK>} and \texttt{<EOS>}
special tokens. Whether the visual advantage persists
at larger scales is an open question. \textbf{Encoder overhead.} The vision
encoder adds modest computation relative to a token embedding lookup, which
would need management at production scale. \textbf{Cross-script comparison.}
The English experiment uses a naive rendering strategy that does not respect
token-length variation, limiting conclusions about alphabetic scripts generally.

\bibliographystyle{plainnat}
\bibliography{full}

\end{document}